# Topic Classification on Spoken Documents Using Deep Acoustic and Linguistic Features

*Tan Liu, Wu Guo, Bin Gu*

National Engineering Laboratory for Speech and Language Information Processing,
University of Science and Technology of China, Hefei, China

`{liutan, bin2801}@mail.ustc.edu.cn, guowu@ustc.edu.cn`

## Abstract

Topic classification systems on spoken documents usually consist of two modules: an automatic speech recognition (ASR) module to convert speech into text and a text topic classification (TTC) module to predict the topic class from the decoded text. In this paper, instead of using the ASR transcripts, the fusion of deep acoustic and linguistic features is used for topic classification on spoken documents. More specifically, a conventional CTC-based acoustic model (AM) using phonemes as output units is first trained, and the outputs of the layer before the linear phoneme classifier in the trained AM are used as the deep acoustic features of spoken documents. Furthermore, these deep acoustic features are fed to a phoneme-to-word (P2W) module to obtain deep linguistic features. Finally, a local multi-head attention module is proposed to fuse these two types of deep features for topic classification. Experiments conducted on a subset selected from Switchboard corpus show that our proposed framework outperforms the conventional ASR+TTC systems and achieves a 3.13% improvement in ACC.

**Index Terms**: topic classification, speech recognition, local multi-head attention

## 1. Introduction

With the rapid expansion of audio media sources such as meeting recordings, broadcast news and voicemail, there is an increasing demand for automatic classification of spoken documents, which makes topic classification (TC) on spoken documents a significant role in information retrieval (IR).

Conventional TC systems on spoken documents are usually designed as pipeline structures, which first transform speech into text through an automatic speech recognition (ASR) module and then perform topic classification on the recognized text through a text topic classification (TTC) module. Regarding ASR, end-to-end models have become popular alternatives to conventional deep neural network-hidden Markov model (DNN-HMM) hybrids because of their simpler model architecture and comparable or even better performance [1-6]. One of the most representative end-to-end models is the connection temporal classification (CTC)-based framework [6]. Moreover, impressive progress has also been made on text topic classification (TTC) with the introduction of DNNs in this field. For example, [7] and [8] employed convolutional neural networks (CNNs) to capture n-gram features at different positions of a sentence for text classification. Recently, BERT (Bidirectional Encoder Representations from Transformers) [9] has revolutionized natural language processing (NLP) research and created state-of-the-art models for a wide range of NLP tasks, including topic classification on text documents. BERT applies a masked language model and next sentence prediction during the pre-training stage and can be finetuned with just one additional output layer for a specific task.

As for the conventional ASR+TTC systems, errors inevitably exist in ASR transcripts, especially in noisy outdoor environments, which are propagated to the TTC module and may affect its performance. Recently, alternative approaches which directly infer semantic information from acoustic features have gained popularity. For example, [10] directly maps the acoustic features of a spoken utterance to intents through an encoder-decoder architecture, which has achieved promising results on spoken language understanding (SLU) task. In addition, [11] pre-trains the deep feature extractor with CTC and then uses layers of convolution and max pooling to arrive at a fixed-dimensional vector for intent classification. However, the above-mentioned works focus on extracting meaning only from acoustic features, while the linguistic context information is always ignored. As we know, linguistic context information plays an important role in SLU. It is a challenge to integrate the linguistic context information in deep feature extraction.

In this work, we also conduct topic classification on spoken documents using deep features directly inferred from acoustic signals. A phoneme-based acoustic model is first trained with CTC loss, which is used to extract deep acoustic features (DAFs). DAFs are the output of the layer before the linear phoneme classifier in the acoustic model, which contains higher-level acoustic information than the acoustic features. Then, these DAFs are converted to deep linguistic features (DLFs) through a phoneme-to-word (P2W) module that is also trained with CTC loss function using the DAFs as input and word labels as targets, and these DLFs contain linguistic context information. Finally, we utilize a local multi-head attention (LMHA) module to fuse these two types of deep features, and the fused features which take both acoustic information and linguistic context information into consider are fed to the final topic classification module. To evaluate the effectiveness of the proposed method, we conduct experiments on a subset selected from switchboard corpus which has 30 topics in total. The proposed system based on the fused deep features outperforms the conventional ASR+TTC systems with a 3.13% improvement in ACC.

The remainder of this paper is organized as follows. In Section 2, we briefly introduce the conventional ASR+TTC approaches. Section 3 provides a detailed description of the proposed approach. Section 4 presents the experimental setup and results. Finally, the discussion and conclusion are presented in Section 5.

## 2. Topic Classification based on ASR+TTC

The ASR+TTC pipeline structure is the mainstream framework for topic classification on spoken documents. The ASR module first converts the speech into text. In this work, the acoustic model of the ASR module is a phoneme-based network trained with CTC loss, which has the same architecture as [13].

Given the decoded word sequence $\widehat{W}$ output by the ASR module, the TTC module predicts the most likely topic class $\hat{C}$ of the spoken document.

$$\hat{C} = \underset{C}{argmax}\, P(C|\widehat{W}) \quad (1)$$

We adopt two popular models, TextCNN [8] and BERT [9], as the TTC module.

## 3. Proposed method

The overall architecture of the proposed method is shown in Fig.1, it includes four modules: an acoustic-to-phoneme (A2P) module used to extract deep acoustic features, a P2W module used to extract deep linguistic features, a local multi-head attention (LMHA) module used to fuse the two kinds of deep features and a topic classification module used to predict the topic using the fused deep features. The following subsections give detailed descriptions of each module.

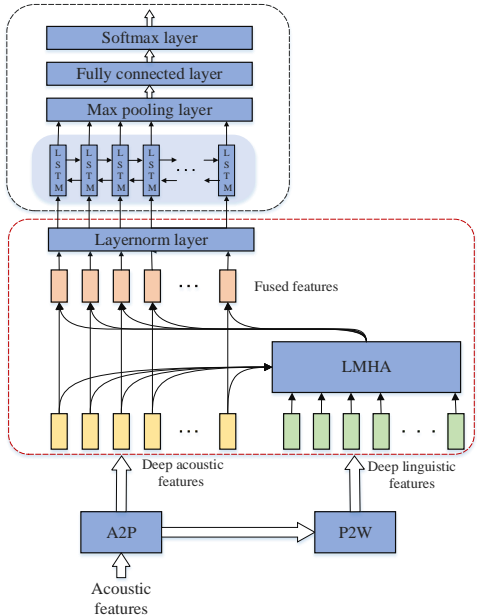

Figure 1: *The overall architecture of the proposed framework*

### 3.1. A2P module

A2P is actually a phoneme-based acoustic model (AM) in ASR systems. Given acoustic feature $X$, the posterior probability of phoneme sequence $P$ can be obtained by:

$$P(P|X) = A2P(X) \quad (2)$$

For conventional ASR tasks, a lexicon and a language model are usually required for decoding. However, in our approach, decoding is never performed and the A2P module is only used to extract deep acoustic features (DAFs) for an utterance. The A2P model consists of a CNN block, a 3-layer bidirectional Long Short Term Memory network (biLSTM) [14] and a linear phoneme classifier. With the trained A2P model, we can extract the deep acoustic features that are the output of the last biLSTM layer. Compared with acoustic features, DAFs contain higher-level acoustic information.

Since there is a 'blank' unit in the CTC-based acoustic model and a lot of frames of an utterance will be recognized as 'blank' by the linear phoneme classifier, the DAFs of these frames will be abandoned in the following process. Through this operation, we can achieve computing efficiency for the following process.

### 3.2. P2W module

P2W module is trained using the CTC loss function with DAFs as input and words as output targets. In other words, the P2W module takes the DAF sequence $\widetilde{p}$ as input to predict the posterior probability of word sequence $W$ during training stage.

$$P(W|\widetilde{p}) = P2W(\widetilde{p}) \quad (3)$$

The P2W module consists of a 2-layer biLSTM followed by a linear word classifier. The deep linguistic features (DLFs) are the output of the last biLSTM layer. Compared with DAFs, DLFs contain linguistic context information which is beneficial for topic classification. Since the model is trained with CTC loss, a lot of frames will be recognized as 'blank'. However, we retain these frames and extract the DLFs for all frames, and there exist well-localised alignment in time axis between the DAF sequence and the DLF sequence, as depicted in Fig.2.

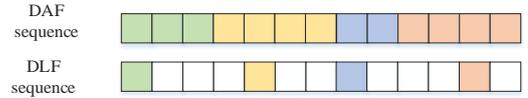

Figure 2: *An example of the well-localised alignment between DAF sequence and DLF sequence. In the DLF sequence, the colored and white boxes represent the DLFs extracted from the frames that are recognized as 'word' and 'blank' by the linear word classifier, respectively. As the figure shows, each DAF has a corresponding DLF (the same color box) which is located nearby it in time axis.*

### 3.3. LMHA

Since DAF sequence $\widetilde{p}$ contains high-level acoustic information and DLF sequence $w$ contains linguistic context clues, we take advantage of the complementary characteristics of $\widetilde{p}$ and $w$ for topic classification on spoken documents. Inspired by [16] and [17], we utilize multi-head attention (MHA) [18] to fuse the two sequences. The architecture of MHA is depicted in Fig.3. In each head of MHA, the query, key and value are first linearly projected to $Q_i$, $K_i$ and $V_i$, where $i$ represents the $i$th attention head. Then, a scaled dot product attention layer computes the weighted sum of $V_i$, which is given as:

$$A_i = softmax(\frac{Q_i K_i^T}{\sqrt{d_k}}) V_i \quad (4)$$

Finally, the outputs of all attention heads are concatenated and once again projected, resulting in the final value, which is calculated by:

$$A = Concat(A_1, A_2, \cdots A_M) W^O \quad (5)$$

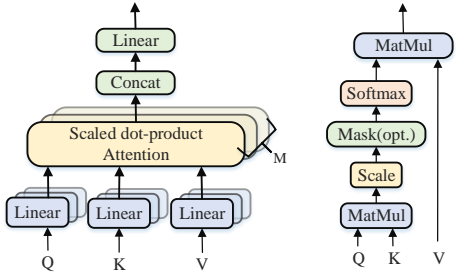

Figure 3: *The left figure is the architecture of multi-head attention and the right figure is the architecture of scaled dot-product attention.*

Since the MHA is expected to learn the alignment between $\tilde{p}$ and $w$ at time axis, which makes each DAF pay more attention to its corresponding DLF, we take $\tilde{p}$ as the query, $w$ as the key and value, such that

$$Q_i = \tilde{p}W_i^Q, K_i = wW_i^K, V_i = wW_i^V \quad (6)$$

where $W_i^Q$, $W_i^K$ and $W_i^V$ are the parameter matrices of the linear layer in the $ith$ head.

As mentioned above, there exist well-localised alignment at time axis between $\tilde{p}$ and $w$, which can directly exclude irrelevant DLF in $w$ to improve the performance. Therefore we adopt local multi-head attention (LMHA) to fuse the two sequences. As for a DAF at position $t$, LMHA restricts the attention region on the DLF sequence to a local scope with a fixed size context window centred at position $t$. For the $ith$ attention head, Eq.4. can be rewritten as.

$$\xi_i = \frac{Q_i K_i^T}{\sqrt{d_k}} \quad (7)$$

$$A_i = softmax(B(\xi_i))V_i \quad (8)$$

$$B(\xi_i) = \begin{cases} \xi_{i_{t,u}} & t - L/2 \leq u \leq t + L/2 \\ -\infty & otherwise \end{cases} \quad (9)$$

where $\xi_i$ represents the attention energy matrix of the DAF sequence in the $ith$ head and $\xi_{i_{t,u}}$ denotes the attention energy between the $tth$ DAF in the DAF sequence and the $uth$ DLF in the DLF sequence. $L$ is the size of local context window, and we set $L$=10 in this work.

Given the DAF sequence $\tilde{p}$ and the output of the LMHA $A$, we apply an element-wise summation of the two sequence to obtain fused deep feature sequence, which is given as:

$$f_a = \tilde{p} \oplus A \quad (10)$$

where $\oplus$ is an element-wise summation operator, $f_a$ is the fused deep feature sequence. Moreover, the two sequences can be concatenated at the feature dimension to obtain fused deep feature sequence $f_c$, as described in Eq.11:

$$f_c = [\tilde{p}, A] \quad (11)$$

where $[,]$ is the a concatenation operator at the feature dimension.

Finally, we apply layer normalization on the fused deep feature sequence to accelerate the convergence.

### 3.4. Topic Classification Module

After obtaining the layernormed fused deep feature sequences, we feed them to a topic classification module which is shown by the black dashed box in Fig.1. In the topic classification module, a 1-layer biLSTM is first used to model the sequential structure of the fused deep features, and then a maxpooling layer is applied over all hidden states in the sequence to construct fixed-dimension representation vectors for spoken documents. These representation vectors will be passed to a subsequent fully connected layer and a softmax layer to predict the topics of spoken documents.

## 4. Experiment

We carry out our experiments on two datasets: Fisher English corpus and Switchboard telephone speech corpus. Both Switchboard and Fisher English corpus are collections of recordings from conversational telephone speech, and each conversation in Switchboard corpus involves a preselected topic. For simplicity, only 900 hours of Fisher English data are used to train the ASR, A2P and P2W modules, while all topic classification experiments are carried out on Switchboard corpus. We select a subset selected from the Switchboard corpus to verify the effectiveness of our proposed model. The subset covers a total of 30 topics and each selected conversation is truncated into several short conversations with a duration about of approximately two minutes.. The subset consists of 4305 short conversations, including 3105 conversations in the train set, 400 conversations in the dev set and 800 conversations in the test set. The accuracy (ACC) is adopted as the metric to evaluate the performance [21]. All our experiments are conducted on PyTorch and our models are optimized with Adam.

### 4.1. Topic classification using ASR transcripts

#### 4.1.1. ASR module

As mentioned in Section 2, we adopt the same phoneme-based network as [13] as the acoustic model for ASR. There are 158 monophonemes (including 'blank' unit) used as modelling units. The acoustic feature is 108-dimensional filterbank features (36 filter-bank features, delta coefficients, and delta-delta coefficients) with mean and variance normalization. The network has a 3-layer biLSTM with 512 hidden units in each direction, and we use dropout operation with the dropout rate of 0.3 before each biLSTM layer. A 3-gram word-based language model is used for decoding. We test the ASR module on the selected Switchboard subset which is also used for topic classification. The WER (word error rate) on the subset is 21.0%. Since there are only 900 hours of training data, this result reaches the average performance level.

#### 4.1.2. TTC module

Before model training, text pre-processing operations such as stop words removal, word stemming or word lemmatization are applied to all the recognized documents for better performance. Two models, the TextCNN and BERT, are used as the topic classification module based on the ASR transcripts. For the TextCNN module, all words in a document are first embedded as 300-dimension vectors by the pretrained GloVe embedding [19] and then fed to the TextCNN. The network has three parallel convolutional layers with identical filters number of 256 and different kernel sizes of (2,3,4), and all convolutional layers use ReLU as the activation function. The outputs of the three parallel convolutional layers are concatenated and pooled as a 768-dimension vector that is used for topic classification. The learning rate of the TextCNN is set to 1e-3.

As for the BERT, we adopt the pre-trained BERT model [20] with hidden size of 768, 12 transformer blocks and 12 self-attention heads for better performance. We finetune the BERT using a learning rate of 5e-5.

For comparison, we adopt two types of words as input, the oracle (human) transcripts and the recognized texts from ASR. The experimental results of different TTC systems are listed in Table 1. Not surprisingly, the systems based on the oracle transcripts outperform those based on ASR recognized text. The ASR errors inevitably deteriorated the accuracy. In addition, we can see that the BERT model outperforms the TextCNN on both types of words. We take the BERT based on recognized texts (ACC=81.75%) as the baseline.

Table 1: *The performance of different TTC module*

| Model | ACC(%) |
|---|---|
| **(Oracle transcripts)** | |
| TextCNN | 86.38 |
| BERT | 87.12 |
| **(Recognized texts)** | |
| TextCNN | 79.38 |
| BERT | 81.75 |

**4.2. Topic classification using DAFs and DLFs**

*4.2.1. A2P module and P2W module*

The trained phoneme-based acoustic model in Section 4.1.1 is also used as the A2P module. The DAFs are extracted from acoustic features with the A2P module, and the DAFs with the recognized 'blank' label are abandoned to improve computing efficiency. Since there are 512 hidden units used in each unidirectional LSTM in the A2P model, the dimension of the DAF is 1024.

As for the training of P2W, we adopt 22095 words as the modeling unit (including 'blank') and the number of hidden units in each direction of the biLSTM is 512. Dropout (rate 0.5) is applied to all biLSTM layers. After the P2W is well-trained, we can extract DLFs with the dimension of 1024.

*4.2.2. LMHA and Topic classification module*

After obtaining the DAF and the DLF sequences, we jointly train the LMHA and the topic classification modules. The LMHA module contains 8 attention heads and has an attention dimension of 1024. For the topic classification module, the biLSTM uses a 1-layer biLSTM with 512 hidden units in each direction, and a dropout rate of 0.5 is applied to the biLSTM. The fully-connected layer has 1024 hidden nodes, and the output layer has 30 nodes corresponding to the number of topics. Different learning rates are used in the LMHA and topic classification modules, the LMHA uses a learning rate of 1e-4 while the topic classification module uses a learning rate of 1e-3. Before feeding the DAFs and the DLFs to the LMHA, we set a maximum length $n = 1600$ for both the DAF and the DLF sequences in the experiments by padding short spoken documents and truncating the long ones.

Besides the proposed method (denoted as LMHA_fusion in Table 2), 4 contrastive systems are also built for comparison. Two different fusion strategies, element-wise summation (Eq. 10) and concatenation (Eq. 11), are also investigated for the LMHA_fusion and GMHA_fusion systems.

**DAFs-only:** This system consists of only the topic classification module in Fig.1 and takes only DAFs as input. In this system, the DAFs are directly fed to the topic classification module, which is similar with end-to-to end SLU systems.

**DLFs-only:** The DLFs-only system has a similar architecture to the DAFs-only system but takes only DLFs as input.

**CONC_fusion:** In this system, DAF and DLF sequences are first encoded as two representation vectors by a DAF encoder and a DLF encoder, respectively. Both of which consist of a 1-layer biLSTM with 512 hidden units in each direction and a maxpooling layer. Then the two representation vectors are concatenated to obtain a fused representation vector that is fed into a linear topic classifier.

**GMHA_fusion:** The GMHA system adopts global multi-head attention which is calculated by Eq.4 to fuse the DAFs and DLFs sequence, and all other setups are same as those of the LMHA_ fusion system.

The experimental results of different systems are listed in Table.2. As shown in Table 2, the conventional ASR+TTC baseline (ACC=81.75%) can achieve better performance than the systems with only DAFs or DLFs as inputs. One plausible explanation is that we cannot apply text pre-processing operations such as stop words removal for both the DAFs-only and DLFs-only systems. However, all the fusion systems (the bottom five rows) outperform the baseline. The results show that the proposed LMHA_fusion(add) achieves the best accuracy with a 3.13% improvement in ACC over the conventional ASR+TTC baseline. These results demonstrate the complementary characteristics of DAFs and DLFs. Moreover, we can observe a 2.63% improvement in the ACC of LMHA_fusion(add) over the GMHA_fusion(add), and this demonstrate the power of the local muli-head attention strategy.

Table 2: *The performance on different TC systems without ASR transcript*

| Model | ACC(%) |
|---|---|
| DAFs-only | 80.12 |
| DLFs-only | 75.62 |
| CONC_fusion | 82.00 |
| GMHA_fusion(add) | 82.25 |
| GMHA _fusion(cat) | 82.00 |
| LMHA_fusion(add) | **84.88** |
| LMHA _fusion(cat) | 83.75 |

## 5. Conclusions

In this work, we propose a framework for topic classification on spoken documents without using ASR transcripts. The proposed framework extracts deep acoustic and linguistic features from acoustic signals, and applies local multi-head attention to fuse these two types of deep features for final topic classification. The experimental results on a subset selected from Switchboard corpus demonstrate the effectiveness of the proposed model.

## 6. Acknowledgements

This work was partially funded by the National Natural Science Foundation of China (Grant No. U1836219)